# Time2Vec Transformer for Robust Gesture Recognition from Low-Density sEMG

**Blagoj Hristov\*, Hristijan Gjoreski, Vesna Ojleska Latkoska, Gorjan Nadzinski**

\*All authors are with the University "Ss. Cyril and Methodius", Faculty of Electrical Engineering and Information Technologies, Skopje, North Macedonia. Corresponding author: Blagoj Hristov (e-mail: hristovb@feit.ukim.edu.mk).

**ABSTRACT**

Accurate and responsive myoelectric prosthesis control typically relies on complex, dense multi-sensor arrays, which limits consumer accessibility. This paper presents a novel, data-efficient deep learning framework designed to achieve precise and accurate control using minimal sensor hardware. Leveraging an external dataset of 8 subjects, our approach implements a hybrid Transformer optimized for sparse, two-channel surface electromyography (sEMG). Unlike standard architectures that use fixed positional encodings, we integrate Time2Vec learnable temporal embeddings to capture the stochastic temporal warping inherent in biological signals. Furthermore, we employ a normalized additive fusion strategy that aligns the latent distributions of spatial and temporal features, preventing the destructive interference common in standard implementations. A two-stage curriculum learning protocol is utilized to ensure robust feature extraction despite data scarcity. The proposed architecture achieves a state-of-the-art multi-subject F1-score of 95.7% $\pm$ 0.20% for a 10-class movement set, statistically outperforming both a standard Transformer with fixed encodings and a recurrent CNN-LSTM model. Architectural optimization reveals that a balanced allocation of model capacity between spatial and temporal dimensions yields the highest stability. Furthermore, while direct transfer to a new unseen subject led to poor accuracy due to domain shifts, a rapid calibration protocol utilizing only two trials per gesture recovered performance from 21.0% $\pm$ 2.98% to 96.9% $\pm$ 0.52%. By validating that high-fidelity temporal embeddings can compensate for low spatial resolution, this work challenges the necessity of high-density sensing. The proposed framework offers a robust, cost-effective blueprint for next-generation prosthetic interfaces capable of rapid personalization.

## 1. Introduction

Reliable and intuitive control is the key to achieving effective upper-limb prosthetics. For a device to be accepted as an extension of the human body, it must translate physiological intent into mechanical action with low latency and high precision [1]. Surface electromyography (sEMG) remains the standard for non-invasive capturing of these biological control signals, providing a glimpse into the neuromuscular activation of the residual limb. However, the stochastic, non-stationary nature of sEMG signals presents a significant classification challenge. As prosthetic systems continue to evolve to support higher degrees of freedom, the ability to distinguish between fine, anatomically similar movements becomes the critical bottleneck for device utility.

Historically, myoelectric control has relied on feature engineering, where experts manually extracted time-domain and frequency-domain metrics to feed classifiers like Support Vector Machines (SVM) or other classical machine learning models [2–5]. While computationally efficient, these methods often discard subtle, high-dimensional information inherent in the raw signal [6]. This limitation has driven the shift toward deep learning approaches capable of learning representations directly from raw data. Deep learning approaches, including Convolutional Autoencoders [3] and Convolutional Neural Networks (CNNs) [8–11], have shown great success by treating sEMG signals as pseudo-one-dimensional images, effectively capturing local spatial patterns and amplitude features. However, standard CNNs often struggle to model long-range temporal dependencies unless they are exceedingly deep. Likewise, while Recurrent Neural Networks (RNNs) and LSTMs are designed for temporal sequences [12, 13], their inability to parallelize creates computational bottlenecks, and they remain susceptible to gradient vanishing when processing long, high-frequency sEMG sequences [14].

To address these limitations, we propose a hybrid and lightweight Transformer architecture integrating Time2Vec [15], a learnable vector representation of time, that is optimized for sparse two-channel sEMG. Unlike static basis functions, Time2Vec learns periodic activation patterns directly from the data, acting as a learnable bank of sinusoidal basis functions capable of synchronizing with the specific motor unit firing rates of different muscle groups. Furthermore, we investigate three distinct feature integration strategies: concatenation, standard addition, and a novel normalized additive fusion to resolve the scale mismatch between spatial features and temporal embeddings without corrupting signal amplitude. We hypothesize that this normalization aligns the latent distributions of the two modalities, allowing the model to successfully superimpose temporal context onto spatial features without signal degradation, thus maximizing the information density of the input embeddings.

In this work, we rigorously evaluate the effectiveness of learnable temporal embeddings for sEMG classification. To ensure that performance improvements are driven by architectural design rather than raw parameter capacity, we benchmark the proposed Time2Vec model against a comprehensive set of baselines, including standard fixed Positional Encodings (PE), a spatial-only ablation (No-PE), and recurrent networks. Based on this analysis, our key contributions are:

1) Validation of the feasibility of a high-capacity Transformer architecture for robust myoelectric control using only two sEMG channels. By employing a two-stage curriculum and integrating extensive data augmentation with robust feature pre-training, we enable the model to learn fundamental movement patterns that generalize across different individuals. While the initial results show that direct transfer to new unseen subjects is ineffective due to domain shifts, we validate a rapid calibration protocol that recovers performance using minimal data from the new user. This confirms the proposed system's utility as a robust feature extractor capable of rapid personalization for next-generation prosthetic interfaces.

2) A comprehensive, iso-dimensional comparison between learnable temporal embeddings (Time2Vec) and established sequence modeling baselines within a Transformer-based sEMG classifier. We benchmark the proposed architecture against three distinct variants: a standard Transformer with fixed sinusoidal encodings, a spatial-only Transformer ablated of positional information (No-PE), and a strong recurrent CNN-LSTM baseline. Our analysis demonstrates that the Time2Vec integration significantly outperforms all three baselines, confirming that the global receptive field of self-attention is superior to recurrence when paired with high-fidelity temporal embeddings.

3) The proposal of a novel design guideline for sEMG-based Transformers by analyzing the trade-off between spatial feature extraction and temporal resolution under a fixed computational budget. Our goal is to challenge the prevailing assumption that maximizing feature extraction capacity is inherently optimal, demonstrating instead that for dynamic flexion tasks, temporal phase information is as valuable as spatial amplitude information, and that balanced architectures (via concatenation) or full-capacity fusion (via normalized addition) yield the best classification performance.

The remainder of this paper is organized as follows: Section 2 reviews the related work, discussing the evolution of deep learning in sEMG and critically analyzing the limitations of standard positional encodings in current Transformer-based frameworks. Section 3 outlines the data preprocessing pipeline, detailing the signal segmentation strategies, the subject-independent partitioning protocol, and the stochastic data augmentation techniques used to enhance model robustness. Section 4 presents the proposed Time2Vec-Integrated Transformer architecture, describing the convolutional feature extraction stem, the implementation of learnable temporal embeddings, and the systematic investigation of feature fusion strategies. Section 5 details the experimental validation, covering the architectural optimization under fixed budgets, the two-stage curriculum learning protocol, and the comparative ablation studies against recurrent and fixed-encoding baselines, concluding with an evaluation of the rapid user-adaptation capabilities. Finally, Section 6 summarizes the findings and suggests directions for future work.

## 2. Related Work

The Transformer architecture [16] has recently emerged as a powerful alternative to CNNs and RNNs, revolutionizing sequence modeling in Natural Language Processing (NLP). By relying entirely on self-attention mechanisms, Transformers can process temporal data in parallel, capturing global dependencies across the entire signal window simultaneously. However, unlike RNNs, the Transformer architecture is inherently permutation-invariant, requiring the injection of Positional Encodings (PE) to provide the model with essential temporal context. The standard approach utilizes fixed sinusoidal functions (sine/cosine) with predetermined frequencies to map time indices to vectors. While the efficacy of Transformers has been established in biosignal processing [17–19], existing literature primarily adopts the standard architecture directly from NLP without accounting for the unique spectral properties of electromyography signals. We believe that this direct transfer of NLP methodology introduces two critical flaws when applied to physiological time-series. Firstly, standard implementations add the positional vector element-wise to the feature vector. In language modeling, where embedding spaces are high-dimensional and sparse, this addition preserves semantic information. However, sEMG signal features represent continuous physical quantities (e.g., amplitude, power) and mathematically adding a sine wave to these features corrupts the signal magnitude and alters the feature distribution. Since sEMG amplitude carries critical information regarding muscle activation intensity, this additive interference degrades the separability of dynamic gesture features [20]. Second, fixed encodings assume a linear, uniform progression of time. In contrast, human motor control exhibits significant temporal warping, meaning that the duration and velocity of a gesture vary stochastically between repetitions and subjects [21]. A fixed encoding imposes a rigid temporal grid that cannot adapt to these non-stationary variations, leading to phase misalignment in the latent space.

A recent study [22] has demonstrated an accuracy of 92.8%, validating the efficacy of Transformers for physiological signal decoding with their proposed CT-HGR framework. However, in order to achieve this state-of-the-art accuracy, they use 128-channel high-density sEMG which they treat as spatial images. This reliance on dense electrode configurations imposes significant hardware complexity and cost, making it impractical for consumer wearables. Furthermore, their architecture utilizes standard learnable positional embeddings, which lack specific inductive bias for the periodic and continuous nature of biological signals. Another recent study [23] has proposed TraHGR, a hybrid Transformer framework that achieves 86.18% accuracy on 49 gestures. However, their approach incorporates a 12-electrode sensor array and a complex dual-path architecture to extract sufficient spatial-temporal features. Furthermore, TraHGR also employs standard learnable positional embeddings, which treat physiological signals as generic tokens without an explicit inductive bias for periodicity. Other positional encoding approaches for the inclusion of advanced temporal embeddings have already been explored with EMG data. For instance, [24] achieved 99.7% accuracy in user authentication by fusing a Transformer with Bidirectional LSTMs. However, this performance relied on a 4-channel input configuration and the heavy computational overhead of recurrent layers. Notably, the author explicitly reported that incorporating Time2Vec as a preprocessing step degraded model accuracy compared to standard encodings. We argue that this limitation stems from treating Time2Vec as auxiliary input features concatenated at the input level, rather than as a learnable latent embedding integrated into the attention mechanism itself. In this work our goal is to demonstrate the feasibility of a lightweight Transformer architecture optimized for sparse, two-channel sEMG. While previous high-density approaches rely on rich spatial information to separate classes, we aim to recover the lost spatial resolution by replacing standard embeddings with properly integrated Time2Vec learnable encodings. This allows us to capture critical temporal information and periodic patterns, enabling robust performance on practical gesture sets with minimal hardware complexity.

## 3. Data Preprocessing

To achieve our objective of evaluating low-density sensing, the experimental validation relies on a two-channel sEMG dataset sourced from [25], which uses Delsys DE 2.x series EMG sensors. Data was sampled at 4 kHz from eight healthy volunteers who provided informed consent, in accordance with the ethical standards of the University of Technology, Sydney. As this study is retrospective and relies solely on these previously anonymized public records, it falls outside the scope of research requiring new ethical oversight in the Republic of North Macedonia. Under the national Law on Health Protection and Law on Personal Data Protection, institutional approval is reserved for work involving identifiable subjects, therefore, no additional ethical permit was deemed necessary for this analysis. The acquisition protocol involved monitoring the electrical activity of the primary muscle groups governing finger manipulation: the flexor digitorum profundus (flexion) and the extensor digitorum (extension). This compact sensor arrangement provides the raw signals for training and evaluating the proposed classifier. The dataset consists of ten distinct dynamic movements, visualized in Figure 1. This specific set, was selected to rigorously test the system's ability to resolve the anatomical crosstalk between adjacent digits using only sparse sensor inputs.

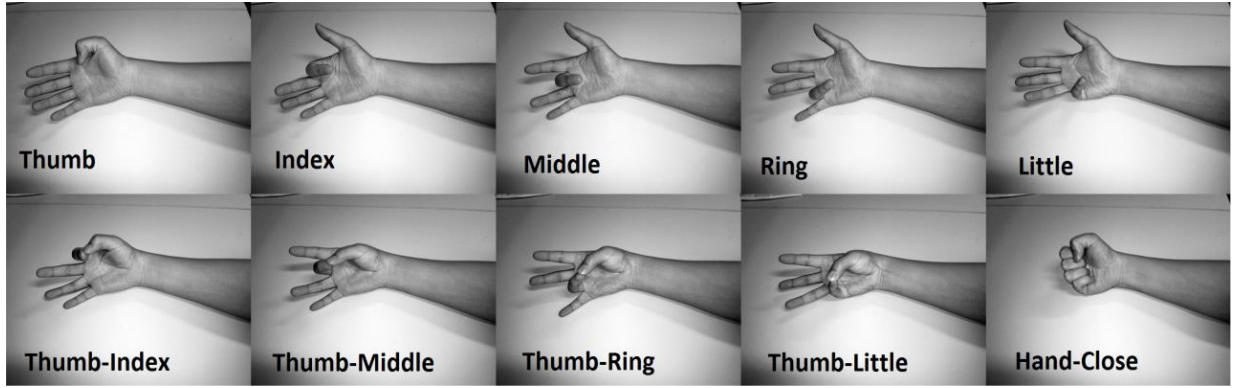

**Figure 1.** Visual representation of the 10-class dynamic gesture set. The experimental protocol includes five individual finger flexions (top row: thumb, index, middle, ring, little), four combined flexions involving thumb opposition (bottom row: thumb-index, thumb-middle, thumb-ring, thumb-little), and a gross hand-close gesture.

### 3.1. Signal Segmentation

The data acquisition protocol involved capturing six independent trials per gesture. Each trial is five seconds long and consists of an isometric hold of the performed gesture. For a prosthetic to feel responsive, the control system needs to react much faster than that, ideally within 300ms, which is roughly the upper limit before a user perceives a delay [26]. To meet this real-time requirement, we limit the signal processing to a 250ms analysis window. We employ an overlapping sliding window technique to partition the continuous 5-second recordings, utilizing a window length of 250ms and a stride of 125ms. This 50% overlap strategy serves two critical functions. First, it effectively doubles the volume of training data while preserving the temporal continuity between sequential states. From an operational perspective, this architecture enables the controller to issue a new motor command every 125ms based on the preceding 250ms of neural activity. This 125ms refresh rate ensures sufficient computational time for actuator response, maintaining a fluid user experience. Second, the classification problem is framed as the identification of intent within these discrete 250ms temporal windows, rather than analysis of the full aggregate trial.

### 3.2. Dataset Partitioning

To guarantee the integrity of our evaluation and strictly prohibit data leakage caused by window overlap, we organized the dataset using a rigorous 8-fold Leave-One-Subject-Out (LOSO) cross-validation protocol, visualized in Figure 2. This strategy ensures that for every iteration, one subject is completely withheld from the initial training process to serve as a novel target user.

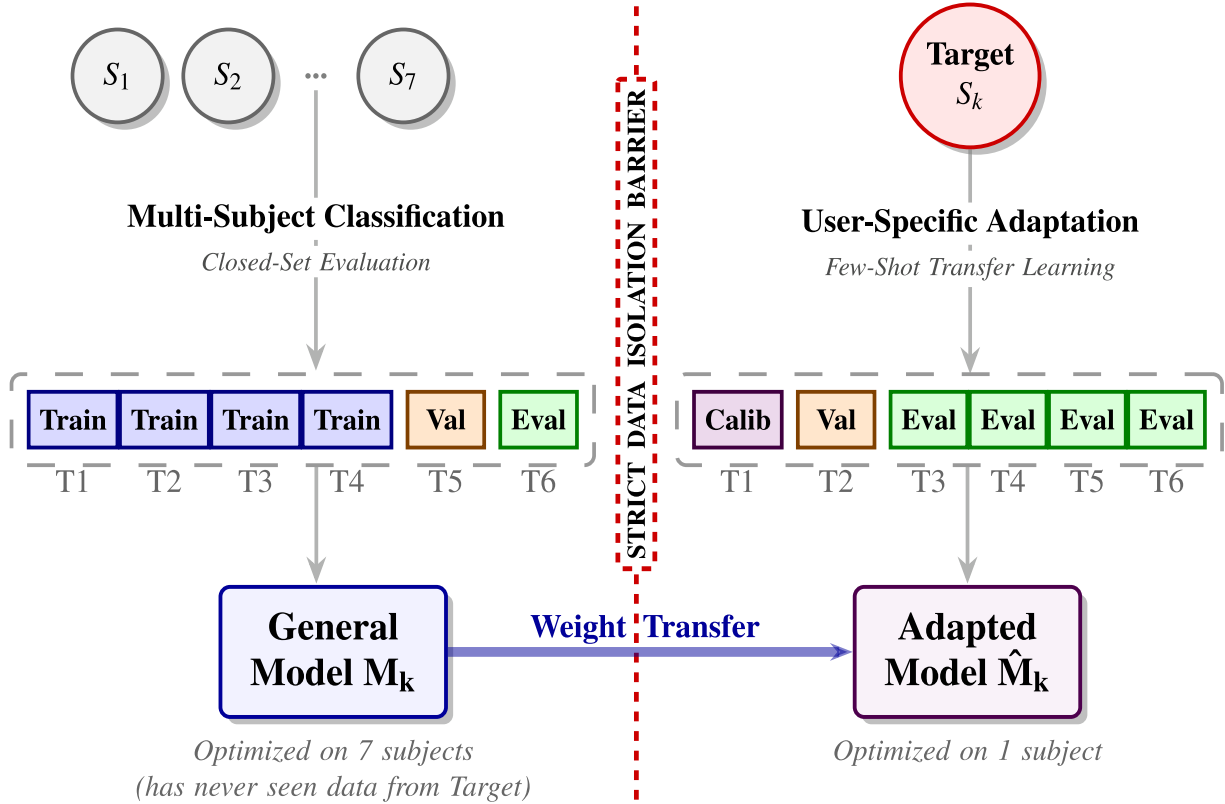

**Figure 2.** Schematic of the cross-validation protocol. To guarantee evaluation integrity, a strict data isolation barrier separates the source cohort from the novel target user. The dataset consists of six recording sessions per subject, denoted as T1 through T6 in the diagram. (Left) Multi-Subject Classification: The model is first optimized on a known cohort of 7 subjects using a chronological split (training: trials 1–4; validation: trial 5; evaluation: trial 6) to learn generalized movement patterns. (Right) User-Specific Adaptation: To test transferability, the pre-trained model is adapted to the single held-out subject who was completely withheld during initial training. A rapid calibration scenario is simulated by using trial 1 for fine-tuning and trial 2 for validation, while trials 3–6 serve as the final unseen test set to measure post-calibration accuracy. This cycle is iterated across all eight folds to ensure robust performance aggregation.

Within each fold, two objectives were analyzed:
1) **Multi-Subject Classification:** The first objective evaluates the model's ability to learn generalized movement patterns from a known cohort of 7 subjects. To assess performance on these known users without overfitting to specific repetitions, we applied a chronological split to the trials of these 7 participants:
    - **Training:** Segments derived from the first four trials were used to optimize model weights.
    - **Validation:** Segments from the fifth trial were isolated for hyperparameter tuning and early stopping.
    - **Evaluation:** The sixth trial was withheld to test the baseline performance on the source subjects.
2) **User-Specific Adaptation:** The second objective evaluates the transferability of the pre-trained model (from the multi-subject classifier) to the single held-out subject. Crucially, the model used here has never seen any data from this subject. To simulate a rapid calibration scenario, the target subject's data was split as follows:
    - **Calibration (Fine-Tuning)/Validation:** The first trial was used to fine-tune the pre-trained model, simulating a short calibration session, while the second one was used for validation.
    - **Evaluation:** The remaining trials (3-6) were used as the final test set to measure post-calibration accuracy.

This cycle was iterated eight times, rotating the held-out subject in each fold. Consequently, the final reported results represent the aggregated mean performance and standard error across all eight folds, ensuring a robust assessment of both the model's baseline learning capability and its adaptability to novel unseen users.

### 3.3. Data Augmentation

Due to the high parameter capacity of the proposed Transformer architecture and the inherent scarcity of data in the two-channel low-density setup, we implemented a comprehensive, stochastic data augmentation pipeline to mitigate overfitting and improve generalization, based on approaches studied in [27]. To effectively double the training set size, we generate exactly one synthetic instance for every original signal window. This augmented counterpart is not limited to a single artifact; instead, it is constructed by passing the original signal through a sequential pipeline where five distinct transformations are applied probabilistically (each transformation has a specific probability of being activated). This ensures that the model is exposed to complex, compound physiological and hardware artifacts without exponentially increasing the dataset size. The transformations, visualized in Figure 3 for one sample from a single channel, include:

- **Gaussian Jittering**: Additive noise was injected to simulate sensor thermal noise and electronic interference.
- **Channel Scaling**: Random multiplicative scaling was applied independently to the two channels to mimic variations in electrode impedance and muscle fatigue levels.
- **Time Warping (Random Resize-Cropping)**: To explicitly address the temporal warping phenomenon described in Section 1, segments were randomly cropped and resized using linear interpolation, effectively simulating stochastic variations in the speed of the gesture execution.
- **Time Masking**: Random temporal segments from the windows were zeroed out to simulate transient sensor disconnection or signal artifacts, forcing the model to rely on global context rather than local spikes.
- **Mix-up:** Finally, convex combinations of pairs of examples and their labels were generated to regularize the decision boundary and encourage linear behavior in-between classes. By introducing this variability, we force the model to look past superficial artifacts like signal noise or speed changes. This ensures it learns the fundamental physiological characteristics of the movement, rather than simply memorizing the specific examples in the dataset.

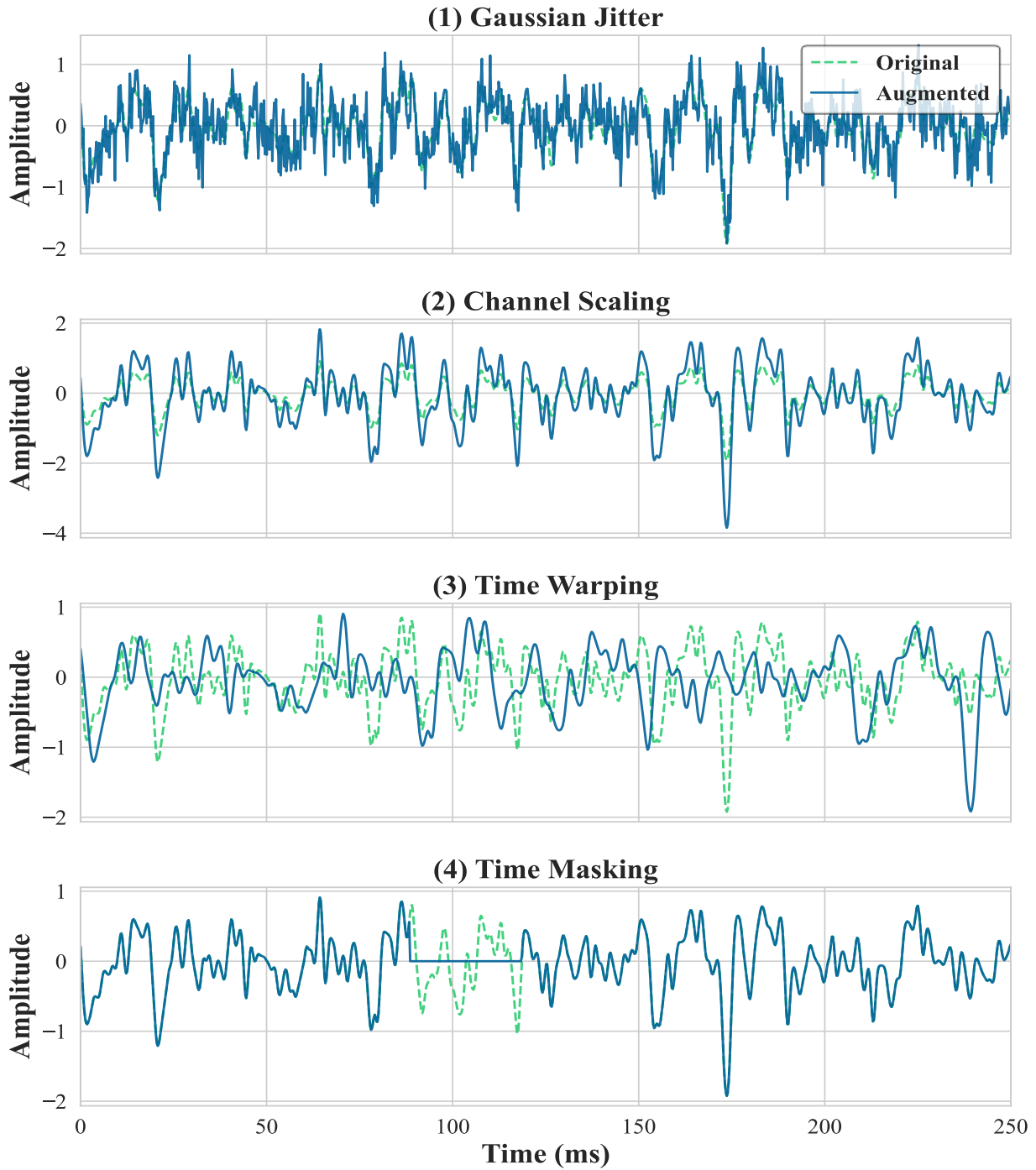

**Figure 3.** Visualization of the stochastic data augmentation pipeline applied to a representative sEMG signal segment (250ms). The dashed green line represents the original, clean signal, while the solid blue line depicts the augmented variation. (1) Gaussian Jitter: Additive noise simulates sensor interference. (2) Channel Scaling: Multiplicative scaling mimics variations in electrode impedance and muscle fatigue. (3) Time Warping: Random resize-cropping introduces non-linear temporal shifts. (4) Time Masking: Random zeroing of temporal segments simulates signal artifacts or sensor disconnection.

## 4. Methods

The proposed framework is designed to facilitate end-to-end classification of dynamic sEMG gestures by learning a direct mapping from raw temporal signals to movement classes. Unlike traditional approaches that rely on manual feature engineering or purely recurrent architectures, our method leverages a hybrid Convolutional-Transformer architecture. This design capitalizes on the complementary strengths of CNNs for local feature extraction and Self-Attention mechanisms for modeling global temporal dependencies. The complete processing pipeline is visualized in Figure 4 and consists of three integrated modules: a convolutional feature extraction stem, a temporal embedding layer, and a Transformer encoder with a classification head.

### 4.1. Convolutional Feature Extractor

While Transformers excel at modeling long-range dependencies, applying self-attention directly to high-frequency raw sEMG samples is computationally prohibitive due to the quadratic complexity of the attention mechanism $[O(N^2)]$. Furthermore, raw sEMG signals are stochastic and noisy, thus attempting to learn global relationships between individual voltage samples is often less effective than analyzing local activation patterns. To address this, our architecture begins with a lightweight CNN stem acting as a trainable tokenizer. The input sequence $X \in \mathbb{R}^{T \times C}$ (where $T = 1000$ represents the 250ms window at a 4kHz sampling rate, and $C = 2$ represents the sensor channels) is processed by two consecutive 1D convolutional blocks. The first layer employs a large kernel size ($k = 61$) to capture broad temporal envelopes, while the second layer ($k = 7$) refines these into sharper, high-level features. Critically, the kernel size of the first layer is deliberately calibrated to create a temporal receptive field of approximately 15.25ms at the 4kHz sampling rate of the data. This window aligns with the physiological duration of a Motor Unit Action Potential (MUAP) [28], allowing the network to capture the complete morphological characteristics of motor unit firings by acting as a learnable matched filter. Preliminary architectural tuning indicated that deviating from this biological prior resulted in performance degradation. Both layers utilize a stride of 2. This progressively downsamples the temporal resolution by a factor of 4 ($T \rightarrow T/4$), compressing the 1000-point raw signal into a dense sequence of approximately 250 feature vectors. This striding operation is critical as it expands the effective receptive field of subsequent layers while significantly reducing the computational load for the Transformer. Following the convolutional operations, the extracted features are projected by a dense layer into a latent feature space of dimension $d_{model}$. This vector, denoted as $Z_{spatial}$, represents the "spatial" information, encoding what type of muscle activity occurred, independent of its specific position in the sequence.

#### 4.1.1. Learnable Temporal Embeddings via Time2Vec

To address the limitations of fixed basis functions, we implement Time2Vec, a learnable vector representation of time. This approach is grounded in the assumption that any time-series signal can be decomposed into a non-periodic (linear) trend and a set of periodic components. Time2Vec defines a temporal embedding $v(\tau)$ for a time index $\tau$ as:

$$v(\tau)[i] = \begin{cases} \omega_i \tau + \phi_i, & if\ i = 0\ (linear\ term) \\ sin(\omega_i \tau + \phi_i), & if\ 1 \leq i < k\ (periodic\ term) \end{cases} \quad (3)$$

where $\omega_i$ (frequency) and $\phi_i$ (phase shift) are learnable parameters optimized during training. Crucially, our implementation utilizes a phase-shifted sine-only formulation for the periodic components. Since, by definition, $cos(\theta) = sin(\theta + \pi/2)$, the learnable phase parameter $\phi$ allows the sine function to approximate any necessary phase alignment, including cosine terms, without redundant parameters. This formulation allows the layer to act effectively as a learnable Fourier Transform. By adjusting the weights $\omega_i$, the model can automatically tune its internal frequency detectors to resonate with the specific firing rates of the underlying motor units, providing a flexible temporal representation that is robust to the natural variability of human movement.

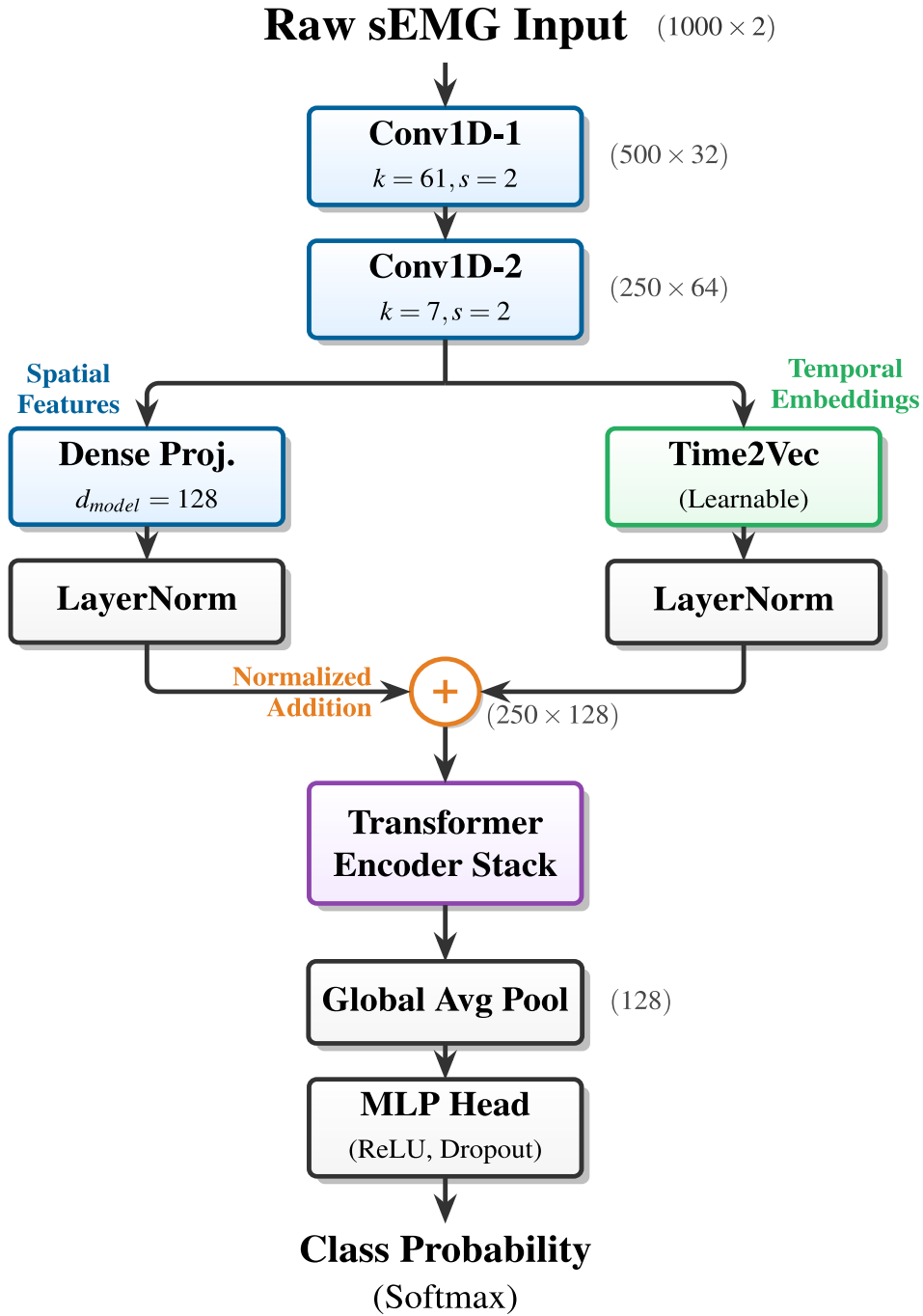

**Figure 4.** Schematic architecture of the proposed Time2Vec-Integrated Transformer. The raw two-channel sEMG input is first processed by a hierarchical convolutional stem to extract local spatio-temporal features and compress the temporal resolution by a factor of 4. The architecture then bifurcates into two parallel pathways: (left) a spatial path, where features are projected to the latent dimension ($d_{model} = 128$); and (right) a temporal path, where Time2Vec learns frequency-adaptive embeddings based on the sequence indices. Then the two paths are fused using a normalized additive mechanism, where both representations are independently layer normalized before element-wise addition, allowing the Transformer encoder stack to process integrated spatio-temporal information without destructive interference. Finally, the sequence is aggregated via global average pooling and classified by an MLP head.

*4.1.2. Feature Fusion Strategies*

We investigate three distinct integration strategies to resolve the challenge of embedding temporal information without corrupting signal amplitude.

- **Standard Additive Fusion:** Typically employed with fixed positional encodings, this method sums the temporal vector element-wise with the spatial feature vector ($Z = Z_{spatial} + Z_{time}$). While computationally efficient, we propose that raw addition introduces destructive interference, where the bounded sine waves of the temporal embedding statistically clash with the continuous amplitude distribution of the sEMG features, potentially degrading the force information vital for classification.
- **Partitioned Concatenation:** To prevent feature interference, we implement a concatenative strategy ($Z = [Z_{spatial}; Z_{time}]$). By appending the temporal vector to the spatial features, we explicitly partition the input space into dedicated subspaces. This guarantees orthogonality between muscle activation and temporal phase, enhancing the resolution of anatomical crosstalk but warranting a reduction in feature capacity to fit within the model dimension budget.
- **Normalized Additive Fusion:** To achieve the benefits of superposition without interference, we implement a normalized additive strategy. We apply layer normalization independently to both the spatial and temporal branches prior to integration ($Z = LN(Z_{spatial}) + LN(Z_{time})$). This aligns the latent distributions of both modalities ($\mu = 0, \sigma = 1$), allowing the model to superimpose temporal context onto spatial features without signal degradation. As demonstrated in our results in Section 5, this strategy yields the best classification performance.

## 4.2. Transformer Encoder

The core of the classification engine is the Transformer encoder, originally proposed by Vaswani et al. [16], which we have adapted specifically for processing a physiological time-series. Unlike RNNs or LSTM networks that process data sequentially, thus inherently limiting parallelization and gradient flow over long sequences, the Transformer utilizes multi-head self-attention to process the entire temporal sequence in parallel. This mechanism allows the model to instantaneously relate the onset of a muscle contraction to its peak amplitude, capturing global trajectory dynamics regardless of the temporal distance between events.

The encoder is composed of a stack of $L$ identical layers. Each layer consists of two primary sub-layers: a multi-head self-attention mechanism and a position-wise Feed-Forward Network (FFN). Crucially, to ensure training stability on stochastic sEMG data, we adopt the Pre-Layer Normalization (Pre-LN) architecture. In this configuration, Layer Normalization (LN) is applied to the input before each sub-layer, rather than after. This modification creates a direct highway for gradient flow through the residual connections, significantly mitigating the vanishing gradient problem during backpropagation and allowing for the robust training of deeper architectures.

*4.2.1. Multi-Head Self-Attention (MHSA)*

The MHSA mechanism enables the model to jointly attend to information from different representation subspaces at different positions. Given an input sequence $H \in \mathbb{R}^{T \times d_{model}}$, the attention function computes a weighted sum of values $V$ based on the compatibility between queries $Q$ and keys $K$. For a single head, these projections are computed as:

$$Q = HW^Q, \quad K = HW^K, \quad V = HW^V, \quad (4)$$

where $W^Q, W^K, W^V \in \mathbb{R}^{d_{model} \times d_k}$ are learnable projection matrices. The scaled dot-product attention is then defined as:

$$Attention(Q, K, V) = \text{softmax}\left(\frac{QK^T}{\sqrt{d_k}}\right)V \quad (5)$$

The scaling factor $\sqrt{d_k}$ is critical for stabilizing the gradients by preventing the dot products from growing large in magnitude, which would otherwise push the softmax function into regions of extremely small gradients. In our multi-head formulation, this operation is performed $h$ times in parallel. The independent attention outputs are concatenated and linearly projected to form the final output:

$$MHSA(H) = Concat(head_1, \dots, head_h)W^O \tag{6}$$

This parallelization allows the model to capture distinct types of temporal dependencies simultaneously, which can lead to, for example, one head focusing on the low-frequency global envelope of the flexion, while another attends to high-frequency transient spikes.

### 4.2.2. Position-wise Feed-Forward Network

Following the attention mechanism, the sequence is processed by a fully connected Feed-Forward Network applied independently and identically to each position. Our implementation deviates from the standard ReLU activation function, instead employing LeakyReLU with a small value for $\alpha$ in order to prevent the dying ReLU problem, ensuring that neurons remain active and gradients continue to flow even for negative activation values inherent in normalized sEMG data. The FFN consists of two linear transformations with a LeakyReLU activation in between:

$$FFN(x) = LeakyReLU(xW_1 + b_1)W_2 + b_2 \tag{7}$$

where $W_1 \in \mathbb{R}^{d_{model} \times d_{ff}}$ expands the dimensionality to project features into a higher-dimensional manifold before compressing them back with $W_2 \in \mathbb{R}^{d_{ff} \times d_{model}}$.

### 4.2.3. Regularization and Stohastic Depth

To prevent overfitting on the limited dataset, we employ a rigorous regularization strategy. In addition to standard Dropout applied to the output of each sub-layer, we implement stochastic depth (DropPath) [29]. This technique randomly drops entire residual branches during training with a probability $p_{drop}$. Mathematically, the output of the $l$-th block becomes:

$$x_l = x_{l-1} + b_l \cdot F(x_{l-1}), \tag{8}$$

where $b_l \in \{0,1\}$ is a Bernoulli random variable with $P(b_l = 1) = 1 - p_{drop}$. This effectively trains an ensemble of shallow networks, forcing the model to learn robust features that do not depend on any single specific layer, thus significantly improving generalization to unseen subjects.

## 5. Experimental Study

The experimental validation of our proposed framework focuses on three primary objectives: determining the optimal allocation of model capacity between spatial and temporal representations, verifying the efficacy of the proposed curriculum learning strategy, and demonstrating the superiority of learnable temporal embeddings in resolving anatomical crosstalk. Consistent with the protocol outlined in Section 3.2, we employed a leave-one-subject-out (LOSO) cross-validation procedure to ensure a rigorous assessment of generalization performance. The reported results are expressed as the mean F1-score aggregated across the eight independent validation folds, providing an unbiased estimate of the classifier's ability to adapt to unseen subjects. Furthermore, to quantify the reliability of these findings, we report the Standard Error (SE) of the mean alongside the F1-scores. This metric characterizes the inter-subject variability, offering a robust statistical measure of the consistency of the model's performance across the study group.

## 5.1. Architecture Optimization

A central idea of this work is that optimal sEMG classification, when implementing the concatenation approach, requires a balanced allocation of model capacity between spatial feature extraction and temporal resolution. To validate this, we first conducted a rigorous ablation study varying the dimension of the Time2Vec embedding ($d_{t2v}$) while keeping the total input budget of the Transformer fixed at 128 dimensions. The results of this analysis are visually presented in Figure 5. Based on these results, we can observe three distinct performance areas:

- **Feature Dominance ($d_{t2v} = 4 - 32$):** In the lower range, the model is heavily biased toward spatial features. While performance generally improves over the baseline (93.7% → 94.4%), the trajectory exhibits high inter-subject variance (indicated by wider error bands), suggesting that insufficient temporal resolution limits the model's stability on complex dynamic gestures.
- **Optimum ($d_{t2v} = 64$):** Classification performance peaks at 95.1% when the capacity is split evenly (64 spatial features vs. 64 temporal features). This balanced allocation confirms that for dynamic flexion tasks, the temporal phase of the signal is as discriminative as its spatial amplitude, and the models benefit most when both are represented equally.
- **Feature Starvation ($d_{t2v} = 96$):** Allocating excessive capacity to time embeddings forces the compression of the CNN feature backbone into just 32 dimensions. This bottleneck causes a performance degradation to 94.6%, validating our fixed-budget experimental design.

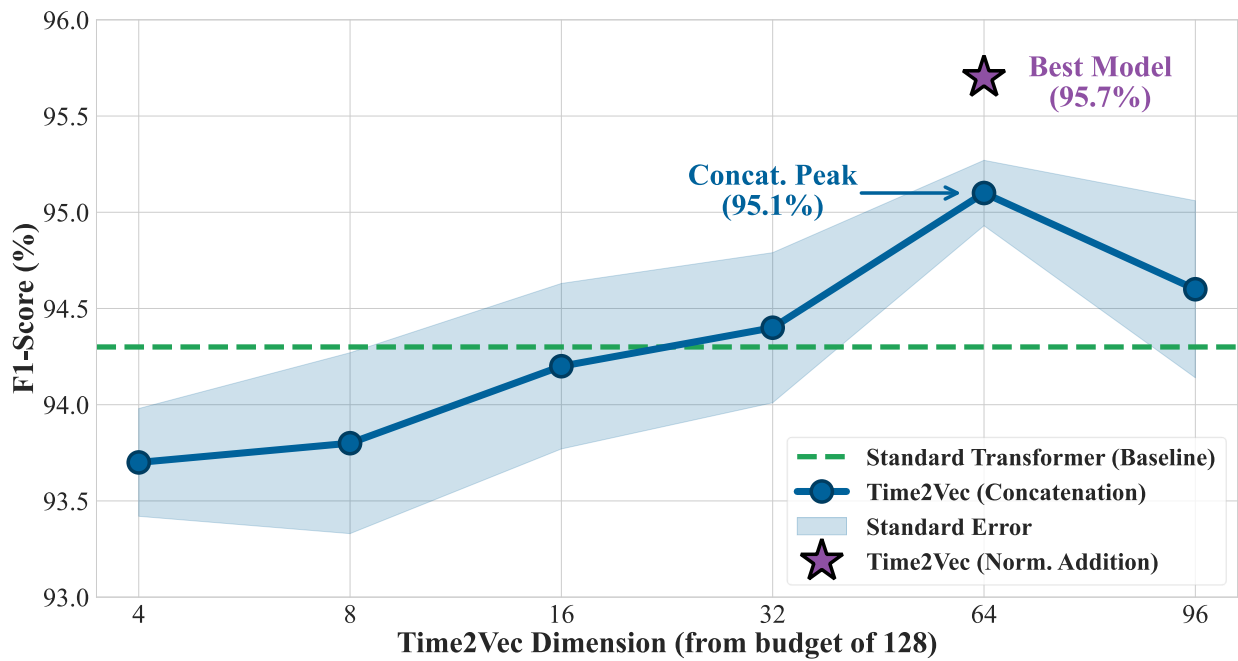

**Figure 5.** Optimization of the temporal embedding dimension ($d_{t2v}$) under a fixed computational budget $d_{model} = 128$. The blue trajectory represents the partitioned concatenation strategy, exhibiting a distinct "inverted-U" curve that peaks at $d_{t2v} = 64$, confirming that a balanced allocation of spatial and temporal capacity is optimal for dynamic flexion tasks. The performance degradation at $d_{t2v} = 96$ highlights "feature starvation", where the CNN feature foundation is excessively compressed. The green dashed line marks the standard Transformer baseline, while the purple star denotes the best performing normalized additive fusion architecture. The blue shaded regions indicate the standard error across the eight cross-validation folds.

## 5.2. Multi-Subject Classification Protocol

Training high-capacity Transformers on low-density sEMG data presents a significant optimization challenge. Direct supervision often leads to overfitting on signal artifacts or convergence to local minima where fine movements remain virtually impossible to distinguish. To navigate this non-convex landscape, we employed a two-stage curriculum learning strategy that prioritizes robust feature extraction before refining decision boundaries.

*5.2.1. Stage I: Robust Feature Pre-Training*

The initial phase of the training procedure focuses on learning a generalized representation of muscle dynamics invariant to sensor noise and speed variations. The model was trained exclusively on the augmented dataset detailed in Section 3.3, using standard categorical cross-entropy (CCE) loss, and employing an Adam optimizer with a high learning rate. This effectively forces the network to learn global waveform signatures rather than memorizing specific time-domain spikes. While this stage yielded a strong generalized model, performance evaluation revealed persistent confusion between anatomically similar gestures (e.g., flexion of the thumb vs. flexion of the index finger), as the CCE gradient was dominated by high-confidence, easily classifiable samples. The performance metrics reported in Table 1 are the average results across all eight folds, comparing the Stage I results when the Time2Vec embeddings are fused with the spatial features through concatenated fusion, simple additive fusion and normalized additive fusion.

**Table 1.** Evaluation of stage I curriculum of multi-subject classification (cross-validation mean F1-score $\pm$ standard error)

| Class | F1-score (Concatenation) | F1-score (Addition) | F1-score (Normalized Addition) |
|---|---|---|---|
| HC | 97.3% $\pm$ 0.23% | 96.9% $\pm$ 0.41% | 96.3% $\pm$ 0.46% |
| T | 91.9% $\pm$ 0.37% | 90.3% $\pm$ 0.58% | 92.1% $\pm$ 0.87% |
| I | 90.5% $\pm$ 0.59% | 89.3% $\pm$ 0.72% | 90.6% $\pm$ 0.68% |
| M | 96.8% $\pm$ 0.39% | 95.0% $\pm$ 0.79% | 96.4% $\pm$ 0.50% |
| R | 92.4% $\pm$ 0.66% | 91.5% $\pm$ 0.68% | 91.0% $\pm$ 0.81% |
| L | 93.5% $\pm$ 0.68% | 90.9% $\pm$ 0.74% | 94.6% $\pm$ 0.85% |
| T-I | 97.6% $\pm$ 0.64% | 97.1% $\pm$ 0.57% | 97.5% $\pm$ 0.66% |
| T-M | 96.6% $\pm$ 0.66% | 96.4% $\pm$ 0.30% | 96.8% $\pm$ 0.70% |
| T-R | 90.1% $\pm$ 0.72% | 90.6% $\pm$ 0.58% | 90.5% $\pm$ 0.83% |
| T-L | 94.0% $\pm$ 0.85% | 93.1% $\pm$ 0.41% | 95.9% $\pm$ 0.41% |
| **Mean** | **94.1% $\pm$ 0.29%** | **93.1% $\pm$ 0.27%** | **94.2% $\pm$ 0.28%** |

In this early stage, the concatenation fusion (94.1% $\pm$ 0.29%) and normalized additive fusion (94.2% $\pm$ 0.28%) perform comparably, with both strategies significantly outperforming the simple additive baseline (93.1% $\pm$ 0.27%). This performance gap is driven by the model's sensitivity to the stochastic noise introduced by data augmentation. In the simple additive approach, the synthetic artifacts in the spatial features are summed element-wise with the temporal embeddings, causing destructive interference that degrades the accuracy of the positional information. In contrast, concatenation mitigates this by maintaining spatial and temporal representations in distinct, orthogonal dimensions, while normalized addition achieves a similar robust effect by stabilizing the feature distributions prior to integration. Both mechanisms effectively protect the learnable temporal cues from being corrupted by noisy, augmented spatial features, thus facilitating faster and more reliable convergence on movement patterns compared to raw addition.

*5.2.2. Stage II: Fine-Tuning*

In the second stage, we shift the training objective from augmentation invariance to distributional precision. We effectively remove the augmentation pipeline to feed the model exclusively with real (non-augmented) source data.

This step addresses the distribution shift caused by the aggressive augmentation in Stage I. While augmentation teaches robustness, it introduces synthetic artifacts that are not present in clean physiological signals. To bridge this gap without unlearning the robust features through catastrophic forgetting, we fine-tune the model on the real source data using a significantly reduced learning rate. This low-magnitude update strategy tunes the decision boundaries to align with the true signal form of the source subjects, ensuring the model is optimized for realistic signal conditions before being tested on the unseen data. The performance metrics reported in Table 2 are the average results across all eight folds, comparing the fine-tuned Stage II results when the Time2Vec embeddings are fused with the spatial features through concatenated fusion (column 1), simple additive fusion (column 2) and normalized additive fusion (column 3).

**Table 2.** Evaluation of stage II curriculum of multi-subject classification (cross-validation mean F1-score $\pm$ standard error)

| Class | F1-score (Concatenation) | F1-score (Addition) | F1-score (Normalized Addition) |
|---|---|---|---|
| HC | 98.1% $\pm$ 0.21% | 97.4% $\pm$ 0.39% | 97.9% $\pm$ 0.33% |
| T | 92.6% $\pm$ 0.61% | 91.1% $\pm$ 0.76% | 92.8% $\pm$ 0.52% |
| I | 91.6% $\pm$ 0.58% | 90.9% $\pm$ 0.86% | 92.1% $\pm$ 0.51% |
| M | 96.5% $\pm$ 0.31% | 96.1% $\pm$ 0.48% | 96.8% $\pm$ 0.39% |
| R | 93.8% $\pm$ 0.55% | 92.5% $\pm$ 0.47% | 94.1% $\pm$ 0.37% |
| L | 95.1% $\pm$ 0.69% | 93.1% $\pm$ 0.28% | 97.0% $\pm$ 0.25% |
| T-I | 98.3% $\pm$ 0.29% | 98.4% $\pm$ 0.30% | 98.4% $\pm$ 0.25% |
| T-M | 96.9% $\pm$ 0.37% | 96.9% $\pm$ 0.33% | 97.9% $\pm$ 0.33% |
| T-R | 92.1% $\pm$ 0.51% | 91.6% $\pm$ 0.50% | 92.8% $\pm$ 0.39% |
| T-L | 95.8% $\pm$ 0.46% | 94.3% $\pm$ 0.42% | 97.0% $\pm$ 0.25% |
| **Mean** | **95.1% $\pm$ 0.17%** | **94.2% $\pm$ 0.23%** | **95.7% $\pm$ 0.20%** |

The refinement stage yields significant gains across all architectures, validating the efficacy of the two-stage curriculum. Notably, the normalized additive fusion strategy significantly surpasses concatenation to achieve the highest overall performance (95.7% $\pm$ 0.20%). This confirms that once the feature distributions are stabilized through layer normalization, utilizing the full shared capacity of the model for both spatial and temporal processing is superior to partitioning the budget. The standard additive fusion remains the weakest performer (94.2% $\pm$ 0.23%), confirming our claim that raw element-wise addition introduces destructive interference, preventing the model from fully resolving the fine temporal nuances of the gesture.

To investigate the root cause of this interference, we visualized the L2-norm distributions of the spatial and temporal feature vectors immediately before the fusion layer. As shown in Figure 6, the standard additive model exhibits a magnitude mismatch where the rigid temporal embeddings clash with the suppressed spatial features in the same numerical range. This lack of separation forces the Transformer to process a mixed signal where temporal position likely drowns out feature content. In contrast, the normalized architecture demonstrates a clear separation of distributions, confirming that the learnable parameters successfully prioritized the informative signal components. This resolved the interference and enabled the superior convergence observed in our metrics.

### 5.3. Comparative Ablation Study

To rigorously quantify the specific architectural contributions of our proposed framework, we benchmarked the best performing Time2Vec implementation (normalized addition) against three distinct models. This comparative analysis is designed to isolate the influence of the feature extraction stem from the temporal modeling mechanism, specifically assessing the efficacy of self-attention compared to traditional recurrence, the baseline discriminative power of spatial features when temporal context is removed, and the specific advantage of learnable, data-driven temporal embeddings over fixed, mathematical formulations. All experimental variants retained the identical CNN feature extraction backbone and were evaluated under the rigorous LOSO cross-validation protocol in order to ensure generalization capability. The performance metrics reported in Table 3 are the average results across all eight folds, comparing the baseline classifier and two ablation models.

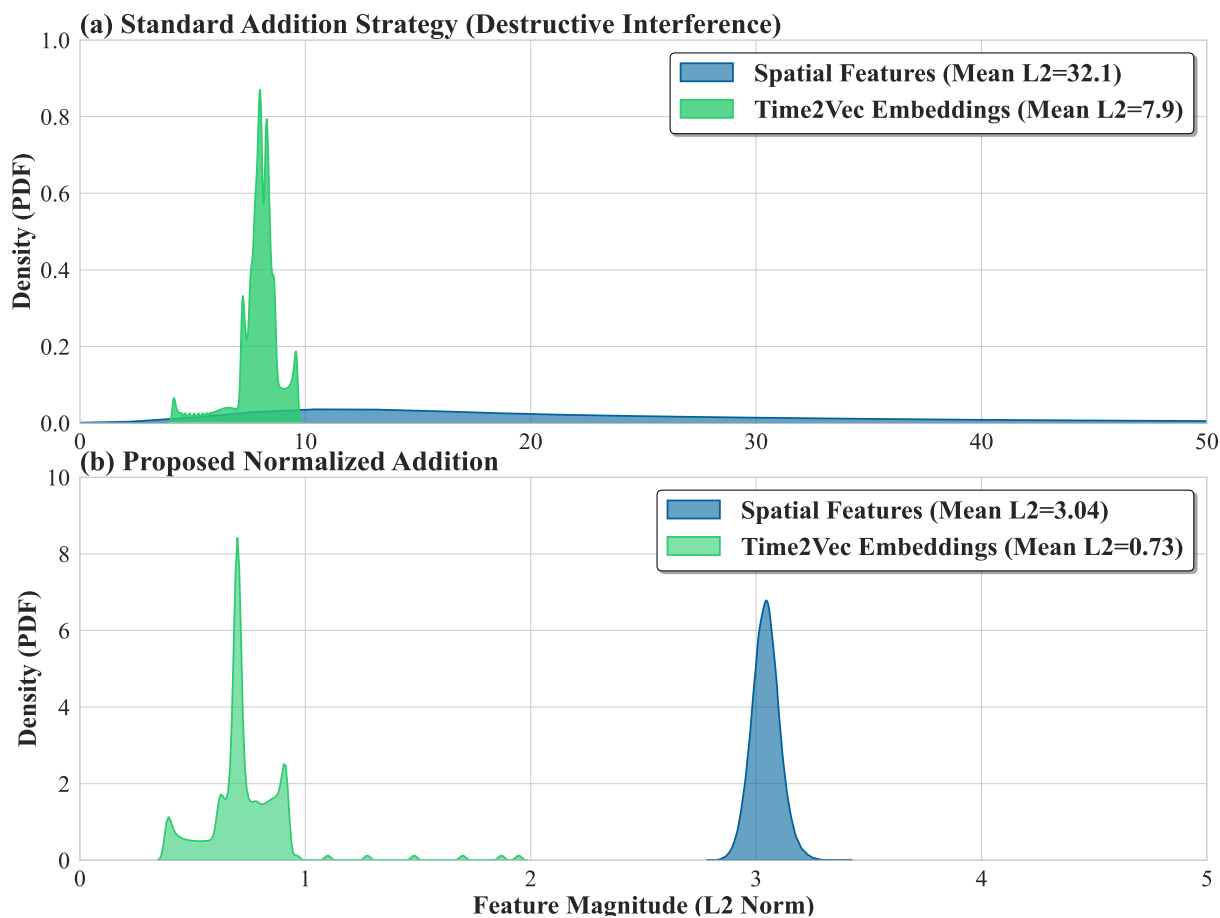

**Figure 6.** Analysis of feature interference. The distribution of L2 norms for spatial features (blue) and Time2Vec temporal embeddings (green) are compared immediately prior to vector addition. (a) Standard Addition Strategy: Without normalization, spatial features exhibit uncontrolled variance with a heavy "tail" extending significantly beyond the visible plot (maximum >500, mean ~32.1), vastly overpowering the bounded temporal embeddings (mean ~7.9). This magnitude mismatch results in destructive interference, where the spatial signal dominates the gradient flow. (b) Proposed Normalized Addition: Applying independent layer normalization prior to addition aligns the feature magnitudes (spatial mean ~3.04, temporal mean ~0.73) to a compatible scale. This alignment ensures both modalities contribute equally to the latent representation.

**Table 3.** Comparative ablation study (cross-validation mean F1-score ± standard error)

| Model | Parameters | Stage I F1-Score | Stage II F1-Score |
|---|---|---|---|
| CNN-LSTM | ~135k | 92.7% ± 0.32% | 93.7% ± 0.30% |
| No-PE Transformer | ~451k | 92.2% ± 0.60% | 93.4% ± 0.45% |
| Standard Transformer | ~579k | 92.9% ± 0.26% | 94.3% ± 0.16% |
| **Time2Vec Transformer** | **~451k** | **94.2% ± 0.28%** | **95.7% ± 0.20%** |

*5.3.1. Baseline Classifier: CNN-LSTM*

To validate the choice of the Transformer architecture, we implemented an established recurrent baseline. While our prior work successfully utilized CNN-LSTMs for this specific task [30], in terms of this comparative study the Transformer encoder was replaced by a scaled CNN-LSTM network in order to match the latent feature dimensionality. This comparison is critical for benchmarking the parallel, global processing capabilities of the Transformer against the sequential, state-dependent processing of traditional RNNs. While LSTMs are the standard for time-series forecasting, they rely on a recurrent inductive bias that processes inputs step-by-step. By comparing this against our model, we test whether the global receptive field of the self-attention mechanism captures long-range dependencies in the sEMG signal more effectively than the gated recurrence of the LSTM. It is important to note that while the CNN-LSTM baseline utilizes fewer parameters (~135k) than the Transformer (~451k), this configuration represents the effective capacity limit for the recurrent baseline on this dataset. Preliminary experiments attempting to scale the LSTM to an iso-parameter arrangement resulted in severe overfitting and convergence instability, a known limitation of recurrent networks when trained on high-variance, stochastic biological signals [14]. Consequently, the performance gap observed here stems from the architectural superiority of the attention mechanism and its ability to utilize capacity effectively, rather than an artificial constriction of the baseline.

*5.3.2. Ablation I: No Temporal Encoding Transformer*

This baseline represents a space-only Transformer where the positional embedding module is entirely removed. In this configuration, the model views the input as a permutation-invariant set of CNN-extracted feature vectors rather than an ordered sequence. By stripping away all explicit temporal indexing, this ablation serves as a negative control. It allows us to measure how much of the classification performance is derived solely from the spatial activation patterns detected by the CNN. If this model performs comparably to the proposed framework, it would imply that the temporal order of muscle activation is irrelevant. In contrast, a significant performance drop here would quantify the distinct value added by modeling the temporal changes in the signal.

*5.3.3. Ablation II: Standard Transformer*

This model implements the standard Transformer architecture utilizing fixed sinusoidal positional encodings added element-wise to the input features. This represents the prevailing state-of-the-art approach in literature for applying Transformers to biological signal processing. The comparison between this fixed approach and our learnable Time2Vec module is central to the goal of this research. While sinusoidal encodings enforce a rigid, predefined temporal grid, they assume that temporal relationships are consistent and periodic in a mathematical sense. By benchmarking against this, we can determine if the irregular nature of sEMG signals is better captured by embeddings that adapt during training, rather than the rigid structure of fixed sinusoidal functions.

The results in Table 3 reveal a distinct performance hierarchy that validates the architectural choices of the proposed framework. The CNN-LSTM establishes a strong recurrent baseline of 93.7% ± 0.30%, demonstrating that sequential processing is effective for capturing temporal dependencies. In contrast, the No-PE Transformer falls below this benchmark at 93.4% ± 0.45%, confirming that a Transformer relying exclusively on spatial activation patterns without any explicit temporal encoding is insufficient for resolving dynamic flexion gestures compared to state-based recurrence. Most importantly, both the standard Transformer (94.3% ± 0.16%) and the proposed Time2Vec architecture (95.7% ± 0.20%) surpass the recurrent baseline. This suggests that the global receptive field of self-attention is superior to LSTM recurrence for this task, given that adequate positional information is supplied. Crucially, this superiority is not a function of model size, seeing as the proposed Time2Vec architecture (~451k) outperforms the standard Transformer (~579k) despite utilizing roughly 22% fewer parameters, confirming that the gain stems from the quality of the learnable embeddings rather than raw capacity. To rigorously validate that this performance gain is systematic, we performed a Wilcoxon signed-rank test on the subject-wise F1-scores. The analysis confirmed a statistically significant difference ($W = 0.0$, $p = 0.008$) between the proposed Time2Vec architecture and the standard Transformer. Notably, the Wilcoxon statistic of zero indicates that the proposed model outperformed the baseline on every individual subject fold, demonstrating that the learnable temporal embeddings offer superior stability across different users compared to fixed sinusoidal encodings. The significant lead of the proposed method further isolates the specific benefit of learnability. While fixed sinusoidal encodings provide necessary temporal context, Time2Vec's adaptive frequency functions better capture the stochastic temporal warping of motor unit firing rates, providing a more precise representation of the movement trajectory.

### 5.4. User-Specific Adaptation

The practical deployment of deep learning prosthetics is fundamentally limited by the high variability of sEMG signals between different individuals. Factors such as unique limb anatomy, varying skin impedance, and distinct muscle recruitment strategies introduce a significant domain shift when moving to a new user. To address this, we designed a user adaptation protocol to explicitly measure this generalization gap and validate the effectiveness of our rapid calibration strategy.

Instead of training a model from the start (which requires a large amount of data) or relying solely on a fixed pre-trained model, we employ a simple data-efficient transfer learning strategy. The experimental protocol evaluated the performance of our optimal pre-trained model (Time2Vec-Integrated Transformer with normalized additive fusion) on a held-out, unseen target subject. This evaluation was conducted in two phases. The pre-trained model was first applied directly to the target subject's data without any modification. This establishes a baseline for the raw transferability of the learned features across the domain gap. We then simulated a short clinical calibration session. The model was fine-tuned using a separate tuning dataset comprising only 2 of the 6 trials from the target subject (approximately 10 seconds of active data). The remaining 4 trials were reserved as the evaluation set. This phase tests the model's learning capability and its ability to personalize decision boundaries to a new physiological signature using minimal data. The results of this procedure are given in Table 4 and they represent the average results across all eight folds where each subject was the chosen test subject once.

The evaluation of the baseline reveals the severity of the domain shift inherent in myoelectric control. When the pre-trained model is applied directly to a novel subject without calibration, classification accuracy collapses to a mean of 21.0%, with complex gestures such as Thumb-Index often failing completely. This drastic performance degradation confirms that the spatial activation maps learned from a source cohort do not generalize to the unique limb geometry and impedance characteristics of a completely new user, rendering a purely zero-shot approach unviable for clinical deployment.

The introduction of a rapid calibration phase utilizing just two training trials per gesture resolves this limitation. Following the fine-tuning protocol, the model's performance recovers dramatically, achieving a mean F1-score of 96.9% ± 0.52%. Surprisingly, this post-calibration accuracy even exceeds the closed-set performance observed in previous experiments, suggesting that the Time2Vec-Integrated Transformer backbone successfully encodes universal dynamic signatures of hand movement during pre-training. Consequently, the system functions as a robust feature extractor that requires only a trivial realignment of its decision boundaries to accommodate the specific physiological profile of a new user, validating the feasibility of a high-performance neuroprosthetic interface incorporating the proposed methodology.

**Table 4.** Evaluation of fine-tuned model (cross-validation mean F1-score ± standard error)

| Class | F1-score (original) | F1-score (fine-tuned) |
|:---:|:---:|:---:|
| HC | 31.4% ± 8.97% | 97.3% ± 0.91% |
| T | 25.3% ± 9.98% | 97.0% ± 1.22% |
| I | 22.3% ± 10.6% | 94.9% ± 1.44% |
| M | 16.5% ± 5.51% | 97.8% ± 0.81% |
| R | 22.4% ± 8.82% | 94.9% ± 1.01% |
| L | 6.5% ± 3.12% | 97.0% ± 1.09% |
| T-I | 2.6% ± 1.09% | 99.4% ± 0.17% |
| T-M | 25.0% ± 9.94% | 97.8% ± 0.61% |
| T-R | 34.5% ± 11.1% | 95.5% ± 1.43% |
| T-L | 23.3% ± 6.79% | 97.8% ± 0.58% |
| **Mean** | **21.0% ± 2.98%** | **96.9% ± 0.52%** |

### 5.5. Discussion

The experimental results presented in this study validate the core idea that a lightweight Transformer, when augmented with learnable temporal embeddings and robust feature fusion, can achieve state-of-the-art myoelectric control performance using minimal sensor inputs. By achieving a mean F1-score of 95.7% ± 0.20% on a complex dynamic gesture set using only two sEMG channels, our framework challenges the prevailing trend in the literature that equates high performance with high-density sensing. This efficiency stands in stark contrast to recent state-of-the-art implementations which largely rely on maximizing spatial resolution to decode intent. For instance, the authors of [22] achieved 91.98% accuracy (rising to 94.8% with offline decomposition) using the CT-HGR framework, however, this performance necessitated a 128-electrode high-density array treating sEMG as 2D images. Similarly, the approach used in [23] achieved between 86% and 93% accuracy with TraHGR using a 12-channel sparse array. By replacing the dense spatial correlations available to these heavier architectures with high-fidelity temporal embeddings via Time2Vec, our framework achieves comparable high-precision performance on complex dynamic gestures using only two channels, demonstrating that high-density arrays are not strictly necessary for decoding fundamental grasp patterns. This suggests that the information density required for robust control can be effectively recovered from the temporal domain if the embedding mechanism is sufficiently adaptive, thus offering a far more viable path for cost-effective consumer prosthetics.

Furthermore, our comparative ablation study explicitly highlights the limitations of the fixed sinusoidal encodings employed by both CT-HGR [22] and TraHGR [23]. While fixed encodings provided a solid baseline (94.3% ± 0.16%), they were consistently outperformed by the Time2Vec architecture, confirming our stipulation that rigid basis functions are suboptimal for physiological signals which exhibit stochastic temporal warping. These results also offer a critical counter-narrative to [24], who reported that Time2Vec degraded accuracy in user authentication tasks compared to BiLSTM-Transformers. We attribute this discrepancy to the integration strategy, whereas the author applied Time2Vec as an auxiliary preprocessing step, we integrated it as a core latent embedding fused with normalized addition. Our findings prove that when temporal features are normalized to match the distribution of spatial features, Time2Vec does not interfere with signal processing but rather synchronizes with the motor unit firing rates, outperforming both standard Transformers and the recurrent BiLSTM/CNN-LSTM baselines utilized in [24].

The inverted-U performance trajectory observed in our optimization study provides a novel design guideline for sEMG Transformers by refuting the assumption that maximizing spatial feature capacity is always optimal. Instead, we demonstrated that a balanced allocation of model capacity yields the highest stability. Additionally, our investigation into fusion strategies successfully resolved the destructive interference problem inherent in standard additive mechanisms. While standard addition plateaued at 94.2% ± 0.23%, our normalized additive fusion strategy successfully aligned the latent distributions of the spatial and temporal branches. This allowed the model to utilize its full parameter budget for joint spatio-temporal processing, significantly outperforming the concatenation strategy (95.1% ± 0.17%), which inherently partitions model capacity.

Beyond classification accuracy, the clinical utility of a myoelectric controller is strictly bounded by its latency. To validate the real-time feasibility of our architecture, we profiled the inference speed on a single core of a standard consumer CPU (AMD Ryzen 9 5900X), simulating a resource-constrained execution environment. The model yielded an average inference processing time of 21.5ms per sample. This latency is highly efficient, consuming only a fraction of the system's 125ms update cycle (defined by the sliding window stride) and ensuring the total system delay remains well below the 300ms threshold where users begin to perceive lag [26]. While embedded microcontrollers operate at lower clock frequencies than workstation CPUs, this low baseline computational cost suggests that the model is well-suited for deployment on edge devices. Furthermore, we anticipate that applying standard optimization techniques, such as int8 quantization or conversion to TensorFlow Lite, would further reduce the memory requirements and execution time, facilitating direct integration into low-power hardware ecosystems.

Finally, the practical utility of the proposed framework is reinforced by the user adaptation experiments. While the poor initial performance of 21.0% ± 2.98% highlights the severity of inter-subject domain shifts inherent in electromyography, the rapid recovery to 96.9% ± 0.52% using only two calibration trials confirms that the model learns robust, transferable dynamic signatures during pre-training. This establishes a practical deployment workflow where the high-capacity Transformer serves as a universal feature extractor that can be personalized to a new user's unique physiology in seconds, effectively bridging the gap between deep learning research and clinical application.

## 6. Conclusion

This work presented a novel, data-efficient deep learning framework designed to enable high-precision myoelectric control using minimal sensor hardware. By integrating learnable Time2Vec embeddings into a lightweight Transformer architecture, we successfully addressed the critical challenge of resolving dynamic gesture kinematics from sparse, two-channel sEMG signals. Our experimental results demonstrate that the proposed architecture achieves a mean F1-score of 95.7% ± 0.20%, significantly outperforming both standard fixed-encoding Transformers and established recurrent baselines (CNN-LSTM). These findings validate the idea that learnable, frequency-adaptive temporal representations are essential for capturing the stochastic temporal warping inherent in biological signals.

Furthermore, this study establishes new design guidelines for the application of attention mechanisms to physiological time-series. We identified normalized additive fusion as the optimal strategy for integrating temporal context, proving that aligning feature distributions prior to addition prevents the destructive interference often observed in standard implementations. Additionally, our capacity optimization analysis refuted the prevailing assumption that maximizing spatial feature extraction is inherently superior. Instead, we demonstrated that a balanced allocation of computational budget between spatial and temporal dimensions yields the highest classification stability for dynamic tasks.

From a practical perspective, the proposed two-stage training curriculum and rapid calibration protocol offer a viable path toward consumer-grade deployment. Although direct transfer remains a fundamental challenge due to intra-subject variability, our system demonstrated the ability to recover with only two calibration trials per gesture. This confirms that the model functions as a robust, universal feature extractor capable of rapid personalization. Future work will focus on deploying this framework onto embedded devices to evaluate real-time latency in closed-loop control scenarios, as well as exploring unsupervised domain adaptation techniques to further reduce the calibration burden for end-users. Finally, by combining high-performance deep learning with low-density sensing, this work offers a practical blueprint for developing more accessible and cost-effective prosthetic systems.

## Data and Code Availability Statement

The dataset used in this study is publicly available from the original authors as cited in [25]. The deep learning models and code used to support the findings of this study are available from the corresponding author upon reasonable request.

## Ethics Approval and Consent to Participate

This study is a retrospective analysis relying exclusively on a publicly available dataset previously collected by Khushaba et al. The original data acquisition involved human participants who provided informed consent in accordance with the ethical standards and approval of the University of Technology, Sydney. As this research utilizes only pre-existing, fully anonymized data records and did not involve any interaction with human subjects or the collection of new identifiable private information, it falls outside the scope of research requiring new ethical oversight under the Law on Health Protection and the Law on Personal Data Protection of the Republic of North Macedonia. Consequently, no additional institutional ethical approval was required for this analysis.

## References

[1] L. C. Smail, C. Neal, C. Wilkins and T. L. Peckham, "Comfort and function remain key factors in upper limb prosthetic abandonment: findings of a scoping review", *Disability and Rehabilitation: Assistive Technology,* vol. 16, pp. 821-830, 2021.

[2] B. Hudgins, P. Parker and R. Scott, "A new strategy for multifunction myoelectric control", *IEEE Transactions on Biomedical Engineering,* vol. 40, no. 1, pp. 82-94, 2002.

[3] B. Hristov, G. Nadzinski, V. O. Latkoska and S. Zlatinov, "Classification of Individual and Combined Finger Flexions Using Machine Learning Approaches", in *2022 IEEE 17th International Conference on Control & Automation (ICCA)*, Naples, Italy, 2022.

[4] P. Kumar et al., "Comparison of Machine Learning Algorithms for EMG based Muscle Function Analysis", in *2022 4th International Conference on Circuits, Control, Communication and Computing (I4C)*, Bangalore, India, 2022.

[5] C. N. Rang et al., "Hand Gesture Recognition using Machine Learning", in *Procedia Computer Science*, 2025.

[6] A. Phinyomark and E. Scheme, "EMG Pattern Recognition in the Era of Big Data and Deep Learning", *Big Data and Cognitive Computing,* vol. 2, no. 3, 2018.

[7] B. Hristov, Z. Hadzi-Velkov, K. H.-V. Saneva, G. Nadzinski and V. O. Latkoska, "Leveraging Convolutional Sparse Autoencoders for Robust Movement Classification from Low-Density sEMG", *arXiv preprint arXiv:2601.23011,* 2026.

[8] X. Chen, Y. Li, R. Hu, R. Hu and X. Chen, "Hand Gesture Recognition based on Surface Electromyography using Convolutional Neural Network with Transfer Learning Method", *IEEE Journal of Biomedical and Health Informatics,* vol. 25, no. 4, pp. 1292 - 1304, 2021.

[9] N. Tsagkas, P. Tsinganos and A. Skodras, "On the Use of Deeper CNNs in Hand Gesture Recognition Based on sEMG Signals", in *2019 10th International Conference on Information, Intelligence, Systems and Applications (IISA)*, Patras, Greece, 2019.

[10] H. Yu, J. Zhang, Y. Gao, H. Wu and X. Ning, "Inverted-Xception: A novel network with multi-channel squeeze-and-excitation temporal deep convolution for high-precision surface electromyography-based gesture recognition", *Knowledge-Based Systems,* vol. 334, 2026.

[11] R. Dermawan, "EMG-Based Hand Gesture Recognition Using Interpretable Deep Learning for Prostheses", *Journal of Electrical Engineering,* vol. 2, 2024.

[12] H. Zhang, H. Qu, L. Teng and C.-Y. Tang, "LSTM-MSA: A Novel Deep Learning Model With Dual-Stage Attention Mechanisms Forearm EMG-Based Hand Gesture Recognition", *IEEE transactions on neural systems and rehabilitation engineering : a publication of the IEEE Engineering in Medicine and Biology Society,* vol. 31, pp. 4749-4759, 2023.

[13] X. Lai et al., "Ratai: recurrent autoencoder with imputation units and temporal attention for multivariate time series imputation", *Artificial Intelligence Review,* vol. 58, 2025.
[14] Z. C. Lipton, J. Berkowitz and C. Elkan, "A Critical Review of Recurrent Neural Networks for Sequence Learning", *arXiv preprint arXiv:1506.00019,* 2015.
[15] S. M. Kazemi, "Time2Vec: Learning a Vector Representation of Time", *arXiv preprint arXiv:1907.05321,* 2019.
[16] A. Vaswani et al., "Attention is all you need", *Advances in Neural Information Processing Systems,* 2017.
[17] M. A. Pfeffer, S. S. H. Ling and J. K. W. Wong, "Exploring the frontier: Transformer-based models in EEG signal analysis for brain-computer interfaces", *Computers in Biology and Medicine,* vol. 178, 2024.
[18] A. Anwar et al., "Transformers in biosignal analysis: A review", *Information Fusion,* vol. 114, 2025.
[19] M. D. Dere and B. Lee, "A Novel Approach to Surface EMG-Based Gesture Classification Using a Vision Transformer Integrated With Convolutive Blind Source Separation", *IEEE Journal of Biomedical and Health Informatics,* vol. 28, 2023.
[20] P. He et al., "DeBERTa: Decoding-enhanced BERT with Disentangled Attention", *arXiv preprint arXiv:2006.03654,* 2021.
[21] E. Scheme and K. Englehart, "Electromyogram pattern recognition for control of powered upper-limb prostheses: State of the art and challenges for clinical use", *Journal of Rehabilitation Research & Development,* vol. 48, pp. 643-660, 2011.
[22] M. Montazerin et al., "Transformer-based hand gesture recognition from instantaneous to fused neural decomposition of high-density EMG signals", *Scientific Reports,* vol. 13, 2023.
[23] S. Zabihi, "TraHGR: Transformer for Hand Gesture Recognition via Electromyography", *IEEE Transactions on Neural Systems and Rehabilitation Engineering,* vol. 31, 2023.
[24] H.-S. Choi, " Feasibility of Transformer Model for User Authentication Using Electromyogram Signals", *Electronics ,* vol. 13, no. 20, 2024.
[25] R. N. Khushaba, M. Takruri, S. Kodagoda and G. Dissanayake, "Toward Improved Control of Prosthetic Fingers Using Surface Electromyogram (EMG) Signals", *Expert Systems with Applications,* pp. vol 39, no. 12, pp. 10731–10738, 2012.
[26] T. R. Farrell and R. F. Weir, "The optimal controller delay for myoelectric prostheses", *IEEE Transactions on neural systems and rehabilitation engineering,* vol. 15, no. 1, pp. 111-118, 2007.
[27] Z. Gao, H. Liu and L. Li, "Data Augmentation for Time-Series Classification: An Extensive Empirical Study and Comprehensive Survey", *Journal of Artificial Intelligence Research,* vol. 83, 2025.
[28] C. J. D. Luca, "Physiology and Mathematics of Myoelectric Signals", *IEEE Transactions on Biomedical Engineering,* Vols. BME-26, no. 6, pp. 312-325, 1979.
[29] G. Huang et al., "Deep Networks with Stochastic Depth", *arXiv preprint arXiv:1603.09382,* 2016.
[30] B. Hristov and G. Nadzinski, "Detection of individual finger flexions using two-channel electromyography", in *XV International Conference ETAI 2021 (Electronics, Telecommunications, Automatics, and Informatics)*, 2021.